%% file: root.tex
\documentclass[letterpaper, 10 pt, conference]{ieeeconf}  

\IEEEoverridecommandlockouts                              

\usepackage{lipsum}
\usepackage{graphicx}
\usepackage[T1]{fontenc}
\usepackage[backend=bibtex]{biblatex} 
\bibliography{references} 
\usepackage{hyperref}
\usepackage{booktabs}
\usepackage{multirow}
\usepackage{pifont}
\usepackage{dblfloatfix}

\usepackage{amssymb}

\usepackage{subcaption}
\usepackage{capt-of}
\usepackage{wrapfig}

\usepackage{tikz}

\usepackage{balance}

\usepackage{float}
\usepackage{makecell}
\usepackage[normalem]{ulem}
\usepackage{booktabs}

\usepackage{amsfonts}
\usepackage{algorithm,algpseudocode}

\usepackage{gensymb}
\usepackage[tbtags]{mathtools}

\usepackage{tikz}
\usetikzlibrary{arrows}
\usetikzlibrary{positioning,calc}
\usetikzlibrary{decorations.pathreplacing}
\usetikzlibrary{decorations.markings}
\usetikzlibrary{fit}
\usetikzlibrary{shapes.callouts}
\usetikzlibrary{shapes.geometric}
\usetikzlibrary{matrix}

\usepackage{textcomp}
\usepackage{siunitx}
\usepackage{multirow}

\usepackage{pgfplotstable}
\usepackage{pgfplots}
\pgfplotsset{compat=1.15}
\usepgfplotslibrary{groupplots}
\usepgfplotslibrary{statistics}

\usepackage[draft,inline]{fixme}
\fxsetup{theme=color}

\title{\LARGE \bf
Accelerating Interactive Human-like Manipulation Learning with GPU-based Simulation and High-quality Demonstrations
}

\author{Malte Mosbach*, Kara Moraw, and Sven Behnke
\thanks{*All authors are with the Autonomous Intelligent Systems (AIS)
Group, Computer Science Institute VI, University of Bonn, Germany;
        {\tt\small mosbach@ais.uni-bonn.de}}%
\thanks{This work has been funded by the German Ministry of Education and Research (BMBF), grant no. 01IS21080, project “Learn2Grasp: Learning Human-like Interactive Grasping based on Visual and Haptic Feedback”.}%
}

\makeatletter
\let\@oldmaketitle\@maketitle
\renewcommand{\@maketitle}{\@oldmaketitle
      \begin{center}
      \includegraphics[width=\linewidth]{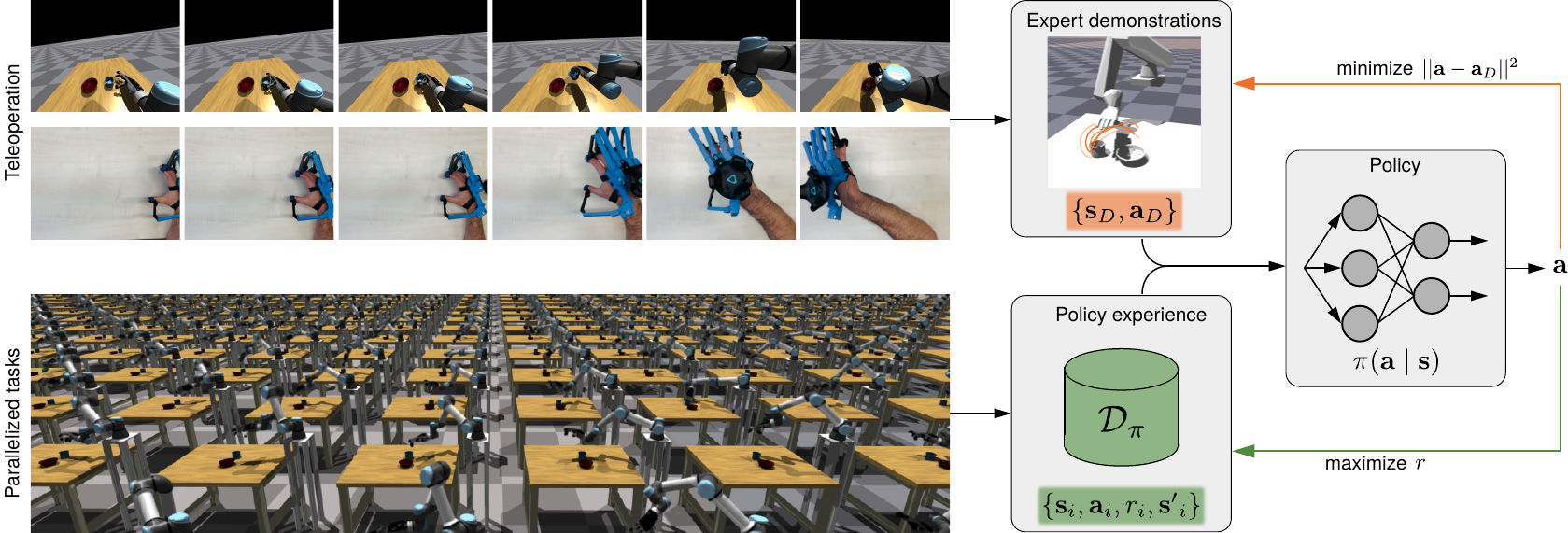}
      \label{overview}
      \captionof{figure}{We combine the strength of massively parallel reinforcement learning to produce proficient continuous control policies with the guidance of human demonstrations to solve challenging manipulation tasks. To this end, we introduce a suite of dexterous  manipulation tasks together with a virtual reality teleoperation framework designed to enable human-like interactive manipulation in contact-rich environments.}
      \end{center}
    }
\makeatother

\begin{document}
\maketitle
\thispagestyle{empty}
\pagestyle{empty}

\begin{abstract}
Dexterous manipulation with anthropomorphic robot hands remains a challenging problem in robotics because of the high-dimensional state and action spaces and complex contacts. Nevertheless, skillful closed-loop manipulation is required to enable humanoid robots to operate in unstructured real-world environments. Reinforcement learning (RL) has traditionally imposed enormous interaction data requirements for optimizing such complex control problems. We introduce a new framework that leverages recent advances in GPU-based simulation along with the strength of imitation learning in guiding policy search towards promising behaviors to make RL training feasible in these domains. To this end, we present an immersive virtual reality teleoperation interface designed for interactive human-like manipulation on contact rich tasks and a suite of manipulation environments inspired by tasks of daily living. Finally, we demonstrate the complementary strengths of massively parallel RL and imitation learning, yielding robust and natural behaviors. Videos of trained policies, our source code, and the collected demonstration datasets are available at  \url{https://maltemosbach.github.io/interactive_human_like_manipulation/}.
\end{abstract}

\section{Introduction}
Anthropomorphic robot hands provide a versatile interface to interact with a human-centric world. The ability to handle various objects and perform dexterous manipulation comes natural to humans but remains an important open problem in robotics. 
Prior work has tackled dexterous manipulation with multi-fingered robot hands through several approaches. Model-based trajectory optimization~\cite{Mordatch2012, Kumar2014} has demonstrated strong performance in simulation, but assumes access to accurate state information and dynamic models. These are difficult to obtain for contact-rich interaction tasks, which makes it challenging to apply these approaches to real-world scenarios and novel objects~\cite{Rajeswaran2017}.
Further, imitation learning may be used, but it relies on high-quality demonstrations to succeed. While these are straightforward to obtain for drone flight or car driving~\cite{Zhang2018}, collecting proficient demonstrations for anthropomorphic robots requires more involved teleoperation setups. Moreover, imitation learning cannot improve upon the observed behaviors.
Alternatively, RL has been used for robot grasping~\cite{Kalashnikov2018, Iqbal2020, Zeng2018, Quillen2018} and object manipulation~\cite{Chen2022b, Rajeswaran2017}, but struggles with the high-dimensional continuous state and action spaces posed by the tasks, leading to extremely high sample complexity. 
So far, manipulation robots remains far from reaching human-level dexterity.

To help closing the gap between robot and human manipulation abilities, we developed a novel framework for learning-based dexterous manipulation, \textit{gym-grasp}, comprising various environments that represent tasks of daily living. 
We integrate two approaches to address the high sample requirements of RL methods on complex manipulation tasks. First, our simulation leverages Isaac Gym~\cite{Makoviychuk2021}, which enables massive parallelization of RL training, resulting in high simulation throughput and empowering the performance of on-policy algorithms. Second, we present a virtual reality (VR) teleoperation framework, designed to enable skillful manipulation in contact-rich environments. In addition to immersive visualization, physical feedback from the environment is a crucial modality humans rely on during interaction with objects~\cite{Huang2007, Richard2021}.
We therefore hypothesize that tactile perception also impacts the quality of demonstrations and integrate haptic feedback based on simulator contact forces using a force-feedback glove. While massively parallel RL has been used to synthesize robust policies for dexterous control, prior works rely on dense rewards to guide policy search towards the desired behaviors~\cite{Chen2022b, Rudin2022}. We demonstrate that imitation learning can be used to make this learning paradigm feasible even for sparse-reward tasks.

The main goal of this work is to foster research in learning-based dexterous manipulation through GPU-accelerated simulation and high-quality demonstrations and evince the complementary strengths of both paradigms. In summary, we present the following contributions:

\begin{itemize}
    \item{\textit{Dexterous Manipulation Benchmark.} We introduce a suite of dexterous manipulation tasks that supports high-performance reinforcement learning via GPU-based physics simulation~\cite{Makoviychuk2021}.}
    \item{\textit{Virtual Reality Teleoperation.} We present a system for immersive VR teleoperation and evaluate the effect of haptic feedback on user preference and task success.}
    \item{\textit{Human Demonstration Datasets.} We provide human demonstration datasets for the considered tasks to enable imitation learning and offline RL for dexterous manipulation.}
    \item{\textit{Combining RL with Demonstrations.} We show the potential benefit of augmenting massively parallel RL with demonstration data.}
\end{itemize}

\section{Related Work}
Robotic grasping and manipulation has been actively studied for decades~\cite{Zhang2022}. Many prior approaches attempt to predict grasp poses either via analytical or data-driven methods~\cite{Kleeberger2020}. Analytical methods are primarily based on mechanics, such as force-closure. They usually assume access to the exact object geometry and friction coefficients at contact points~\cite{Nguyen1988, Markenscoff1989}. Data-driven methods rely on machine learning and some form of training data, but can sometimes dispense with the very high demands of analytical methods for perfect observability of the work space and known object models. Nevertheless, both variants are fundamentally concerned with grasp synthesis, which is the problem of finding a suitable configuration of the robot's end effector, as to grasp a target object. This is in stark contrast to the dexterous manipulation observed in humans, which continuously interleaves perception and action. Recent works utilize RL to achieve such dexterous control. Kalashnikov et al.~\cite{Kalashnikov2018} demonstrate that RL is able to synthesize a vision-based grasping policy that can pick up unseen objects with a parallel gripper, at the cost of hundreds of thousands of real-world grasp attempts. Further, Rajeswaran et al.~\cite{Rajeswaran2017} explore the use of demonstrations for learning dexterous manipulation policies, but do not consider tactile perception. Chen et al.~\cite{Chen2022b} study RL for in-hand reorientation and demonstrate strong performance of parallelized learning in GPU-based simulations.

Using teleoperation to collect demonstrations is an especially suitable paradigm for human-like robots due to the similar morphology and straightforward control mapping. Prior works have used vision-based teleoperation via motion capture data~\cite{Kumar2015} or hand pose estimation~\cite{Qin2021, Qin2022, Handa2020}. While vision-based pose estimation has low hardware requirements, it limits the workspace and cannot provide feedback from the environment. Zhang et al.~\cite{Zhang2018} demonstrate how consumer-grade VR controllers can be used to teleoperate a PR2 robot, but focus neither on anthropomorphic end-effectors, nor on haptic feedback. Kim et al.~\cite{Kim2022} explore force feedback for bilateral teleoperation, but operate with two-finger grippers. Commercially available systems, such as the Shadow Teleoperation System \cite{ShadowTeleoperationSystem2022} provide haptic feedback and could also be used to collect demonstrations in contact-rich domains. However, this requires expensive, specialized hardware. Our system, on the other hand, enables haptic feedback during teleoperation of an anthropomorphic hand based on inexpensive VR-components and can be used with different robotic hands.  

Concurrently to our work, Chen et al.~\cite{Chen2022} introduced a benchmark for bimanual hand manipulation based on Isaac Gym. Unlike our work, their focus is on learning to coordinate two hands and they do not consider learning from human demonstrations.

\renewcommand\thefigure{\arabic{figure}}    
\setcounter{figure}{1}

\begin{figure}[t]
      \centering
      \includegraphics[width=\columnwidth]{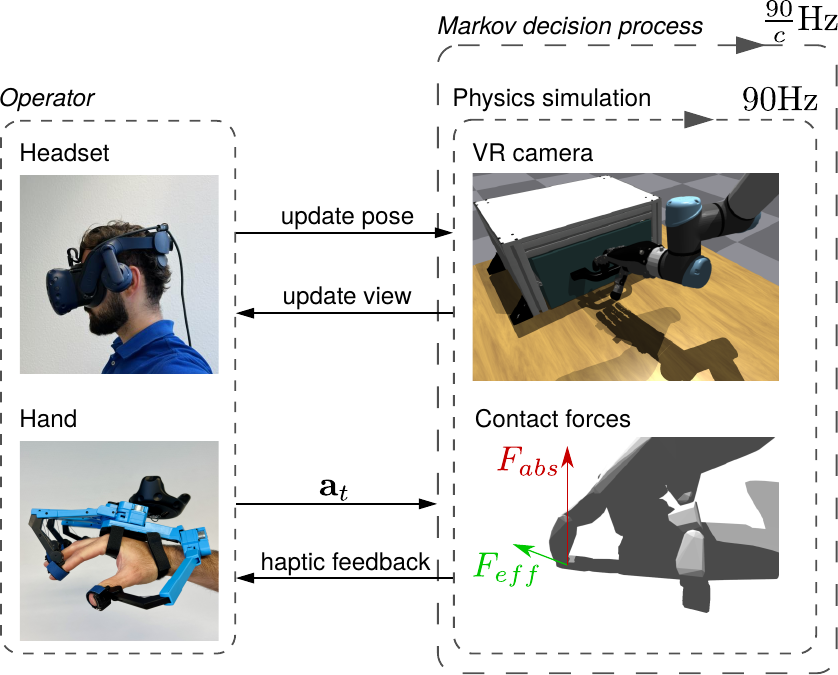}
      \caption{Teleoperation framework overview.}
      \label{teleoperation_framework}
\end{figure}

\section{Virtual Reality Teleoperation}
In the following, we introduce our VR teleoperation system and explain how it facilitates the collection of proficient demonstrations on contact-rich manipulation tasks.

\subsection{System Overview}
Our operator interfaces combines the Vive VR system and the SenseGlove DK1 force-feedback glove. Fig.~\ref{teleoperation_framework} depicts the interaction of the main components. The operator observes the simulated scene through a dedicated camera in Isaac Gym. We update the camera pose instantaneously to follow the operator’s head movements tracked by the headset. For each finger, the SenseGlove detects 4\,DoF finger joint positions and features separately controllable force and vibration feedback. We mount a Vive tracker on top of the glove, providing sub-millimeter 6\,DoF pose information at 90\,Hz. Fusing the information from both devices enables intuitive control of the anthropomorphic robot hand. We employ a clutch-like mechanism, where the user can move their hand freely and only starts teleoperating the robot hand when indicated via the keyboard. The pose of the operator and robot hand then remain locked until the termination of the episode. 

While an update frequency of 90\,Hz is recommended for immersive VR operation, we may want to select actions at a lower rate since RL and imitation learning become increasingly challenging for long horizon tasks. We therefore decouple the update frequency of the head-mounted display and force feedback from the control frequency of the Markov decision process (cf. Fig.~\ref{teleoperation_framework}). Accordingly, we divide the simulation side into the actual physics simulation, which runs at 90 steps per second to provide fast updates for the headset and haptic feedback, and the MDP, which queries actions at $\frac{90}{c}$Hz, where $c$ is the control frequency interval. We found a control frequency of 30 Hz to be sufficient for accurate control by teleoperation and effective learning via RL.

\subsection{Haptic Feedback}
As touch is of crucial importance to humans when performing dexterous manipulations, we incorporate this modality via force and vibration feedback. To allow the user to feel the presence of objects, we determine the rigid-body contact forces of the fingertips and decompose them into the absolute force and a directional component, which acts through the fingertip and prevents the hand from closing. The component of the force acting against closing a finger, $F_{eff}$, is mapped to the SenseGlove's force feedback command, which activates a braking system and increases the resistance to closing the finger further. This mimics the resistance felt when holding an object and is beneficial for grasping and transporting objects. The absolute force magnitude values are smoothed using a simple moving average low-pass filter. We then map the high-frequency components of the absolute force $||F_{abs}|| - \mathrm{MA}(||F_{abs}||)$, where $\mathrm{MA}$ refers to the moving average, to the vibrational feedback. This results in feedback responses for sudden increases in the contact force, such as collisions, but no feedback for sustained contact.

\section{Learning Dexterous Manipulation}

\subsection{Task Design}

\begin{figure}[t]
      \centering
      \includegraphics[width=\columnwidth]{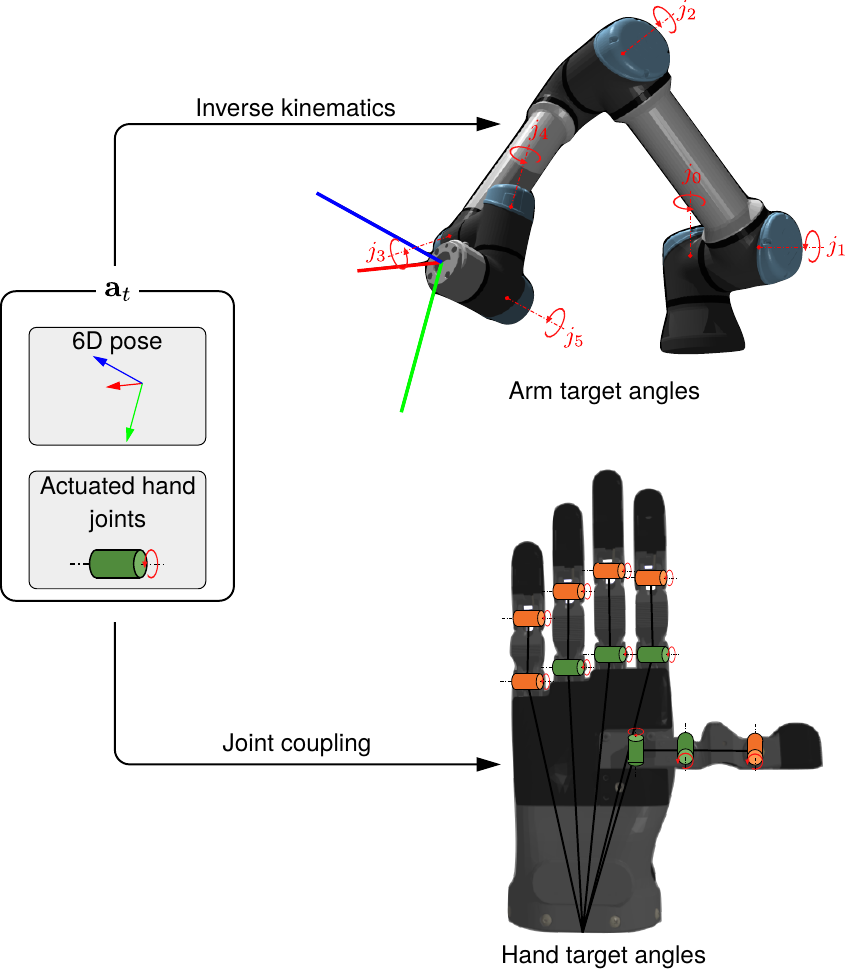}
      \caption{Actuation of robot arm and hand. We actuate the robot hand via the 6D target pose of the wrist and the target joint angles of the fingers. Only the joints of the hand marked in green can be actuated directly. The orange joints are underactuated and positioned via equality constraints.}
      \label{sih_kinematics}
\end{figure}

All implemented tasks feature a robotic arm and hand that we have access to for real-world experiments (in future work). Specifically, a UR5 arm and Schunk SIH hand are used. The actuation of the hand features 5 degrees-of-freedom (DoF). Tendons are used to control the rotating of the thumb and bending of the fingers, resulting in the coupled control scheme depicted in the bottom right of Fig.~\ref{sih_kinematics}, which we replicate through the coupling of position targets in Isaac Gym. The actuation scheme is shown in Fig.~\ref{sih_kinematics}. Actions are 11-dimensional and are interpreted as the relative change to the desired wrist pose (6 dimensions) and finger positions (5\,DoFs of the Schunk SIH hand). We compute the UR5 joint position targets from the desired wrist pose using the Jacobian transpose method. 

\begin{figure}[t]
      \centering
      \begin{subfigure}[b]{0.235\textwidth}
         \centering
         \includegraphics[width=\textwidth]{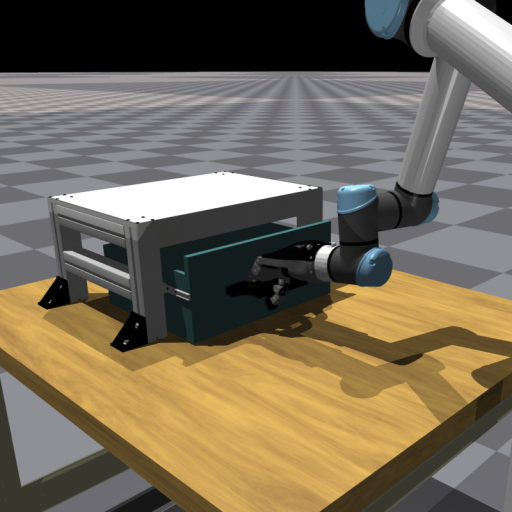}
         \caption{OpenDrawer}
         \label{tasks:OpenDrawer}
     \end{subfigure}
     \hfill
     \begin{subfigure}[b]{0.235\textwidth}
         \centering
         \includegraphics[width=\textwidth]{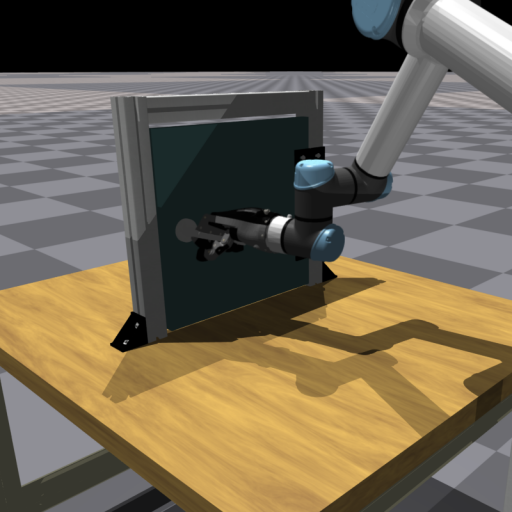}
         \caption{OpenDoor}
         \label{tasks:OpenDoor}
     \end{subfigure}
     
     \vspace{0.2cm}

     \begin{subfigure}[b]{0.235\textwidth}
         \centering
         \includegraphics[width=\textwidth]{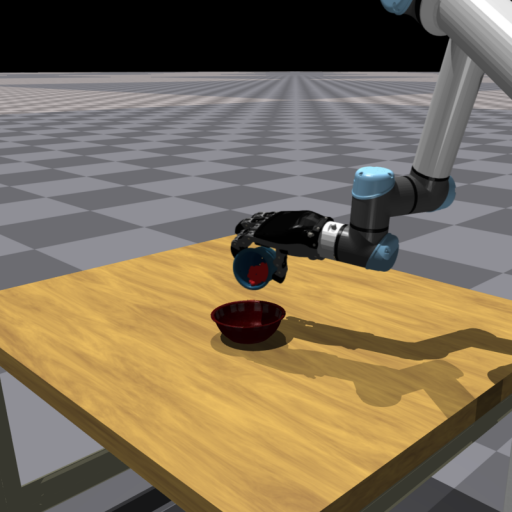}
         \caption{PourCup}
         \label{tasks:PourCup}
     \end{subfigure}
     \hfill
     \begin{subfigure}[b]{0.235\textwidth}
         \centering
         \includegraphics[width=\textwidth]{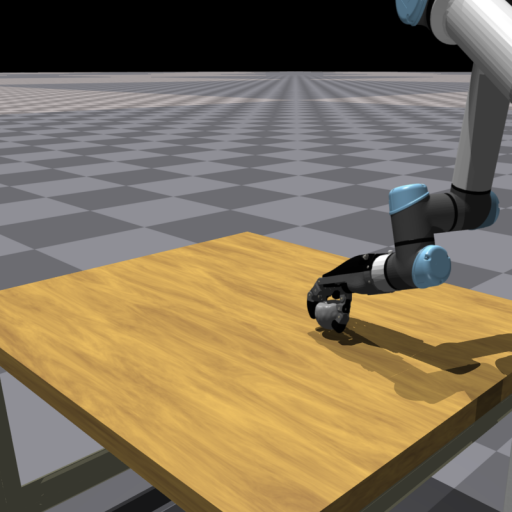}
         \caption{LiftObject}
         \label{tasks:LiftObject}
     \end{subfigure}

      \caption{Dexterous manipulation tasks.}
      \label{tasks}
\end{figure}

The selected manipulation tasks are shown in \ref{tasks}.
All tasks provide proprioceptive observations of wrist pose and finger positions of the robot. On the \textit{OpenDrawer} task, the agent is informed about how far the drawer is opened, from which the handle position can be inferred. Dense rewards penalize the distance of the robot hand from the handle, while continuously rewarding the opening of the drawer. The task is solved once the drawer is pulled open by 0.2m, which causes the environment to terminate. The \textit{OpenDoor} task provides the pose of the door handle and is completed successfully once the door is opened by 45\degree. When dense rewards are used, the agent is rewarded continuously for how far the door is opened and the distance of the robot hand to the door handle is penalized. On the \textit{PourCup} task, we provide the pose of the full cup. The agent is rewarded proportional to the amount of particles poured into the bowl, while the distance of the hand to the cup is penalized. The \textit{LiftObject} task exposes the pose of the object. When using dense rewards, the agent is rewarded for positive changes to the object height, while the distance of the robot hand from the object is penalized. The task is completed successfully and terminates once the object is lifted 0.2m above the table. In addition to the dense reward functions described above, we also investigate sparse rewards because we want to show how imitation learning can be used to bridge the performance gap between the two. Sparse rewards are 0 if a task is completed successfully and -1 otherwise for all tasks.

\subsection{Massively Parallel Reinforcement Learning}
While GPU-based parallelization has been widely adopted for neural network training, data collection in reinforcement learning is still mostly performed via single CPU simulations. Parallelizing data collection by running many instances of a robot in a single simulation has the potential to accelerate RL training by several orders of magnitude~\cite{Makoviychuk2021, Rudin2022}. We leverage the high-performance capabilities of Isaac Gym to scale up our training on dexterous manipulation tasks. To show the efficiency of our framework, we evaluate the simulation throughput during RL training on a single machine with an AMD Ryzen 9 5950X CPU and NVIDIA RTX A6000 GPU. All tasks run 16,384 parallel environments. Note that the number of simulated steps is three times higher than the MDP steps given in Table~\ref{simulation_throughput} because we maintain the control frequency interval of three to obtain consistent results for imitation and reinforcement learning.

While the massive parallelization of agent instances in a single simulation leads to rapid training, it also presents unique challenges. Take the LiftObject task (see Fig.~\ref{tasks:LiftObject}), for example. Since objects of arbitrary geometry may be used, we drop them onto the table from a random seed pose to find feasible initializations before starting the actual task. This procedure cannot simply be repeated when an environment terminates, since a step in the physics simulation to drop the object would also advance the simulation in all other environments. Instead, we only drop the objects at random positions initially and then reset each environment to the exact pose found at the beginning. The randomization of object positions is therefore handled by the large number of environment instances instead of the separate reset phase.

\begin{table}[t]
\caption{Simulation throughput (MDP steps / s).}
\label{simulation_throughput}
\centering
\resizebox{\columnwidth}{!}{%
\begin{tabular}{@{}cccc@{}}\toprule
\multicolumn{1}{c}{OpenDrawer} & \multicolumn{1}{c}{OpenDoor} & \multicolumn{1}{c}{PourCup} & \multicolumn{1}{c}{LiftObject} \\ \midrule

$58,887 \pm 1,726$ & $56,513 \pm 1,442$ & $21,977 \pm 507$  & $58,980 \pm 1,224$ \\
\bottomrule
\end{tabular}}
\end{table}

\subsection{Imitation and Reinforcement Learning}
We aim to demonstrate the complementary strengths of imitation learning and massively parallel RL for dexterous manipulation. RL models the agent-environment interaction as a Markov decision process (MDP). An MDP is defined by the tuple $\mathcal{M} = (\mathcal{S}, \mathcal{A}, R, T, \gamma)$, where $\mathcal{S}$ and $\mathcal{A}$ are the sets of states and actions, $T: \mathcal{S} \times \mathcal{A} \times \mathcal{S} \rightarrow \mathbb{R}_{+}$ is a state-transition probability function, which represents the probability of transitioning to the next state $\mathbf{s}_{t+1} \in \mathcal{S}$ given the current state $\mathbf{s}_{t} \in \mathcal{S}$ and action $\mathbf{a}_{t} \in \mathcal{A}$, $R: \mathcal{S} \times \mathcal{A} \rightarrow \mathbb{R}$ is a reward function, and $0 \leq \gamma \leq 1$ is a discount factor. The state-action marginal of the trajectory distribution induced by a policy $\pi(\mathbf{a}_t | \mathbf{s}_t)$ is denoted as $\rho_\pi (\mathbf{s}_t, \mathbf{a}_t)$. The objective of the RL problem is to maximize the expected sum of rewards discounted over time $J = \sum_t \mathbb{E}_{(\mathbf{s}_t, \mathbf{a}_t) \sim \rho_{\pi}}[\gamma^{t}R(\mathbf{s}_t, \mathbf{a}_t)]$. 

We employ proximal policy optimization (PPO)~\cite{Schulman2017}, which is a state-of-the-art on-policy RL algorithm and base our implementation on~\cite{rl-games2022}. For imitation learning, we use simple behavior cloning (BC) and leverage the implementation provided by Robomimic et al.~\cite{Robomimic2021}. Hence, we collect demonstration data as state-action tuples $(\mathbf{s}_i, \mathbf{a}_i)$ and train the policy to reproduce the expert actions for the same states. Finally, we combine reinforcement and imitation learning by incorporating the losses of both paradigms to form a demo augmented policy gradient (DAPG) algorithm~\cite{Rajeswaran2017}. In contrast to the original variant, which employs a natural policy gradient, we stick with the now widely used PPO algorithm. The trade-off between both losses is weighed as $\mathcal{L}_{DAPG} = \mathcal{L}_{PPO} + \lambda_0 \lambda_1^{k} \mathcal{L}_{BC}$, where $k$ is the current training epoch. We set $\lambda_0 = 50$ and $\lambda_1=0.99$ in all experiments.

\section{Experiments}
We investigate the effectiveness of learning from demonstrations collected by our teleoperation framework, as well as the potential of utilizing demonstration data in massively parallel on-policy RL. Therefore, we aim to address the following questions:

\begin{enumerate}
    \item Can the demonstrations collected with our VR teleoperation system be used to train successful policies for dexterous manipulation tasks?
    \item How does learning from scratch in the massively parallel regime perform on the selected tasks?
    \item How does incorporating demonstration data into the learning process impact the performance of on-policy RL?
\end{enumerate}

\begin{figure}[t]
      \centering
      \includegraphics[width=0.75\columnwidth]{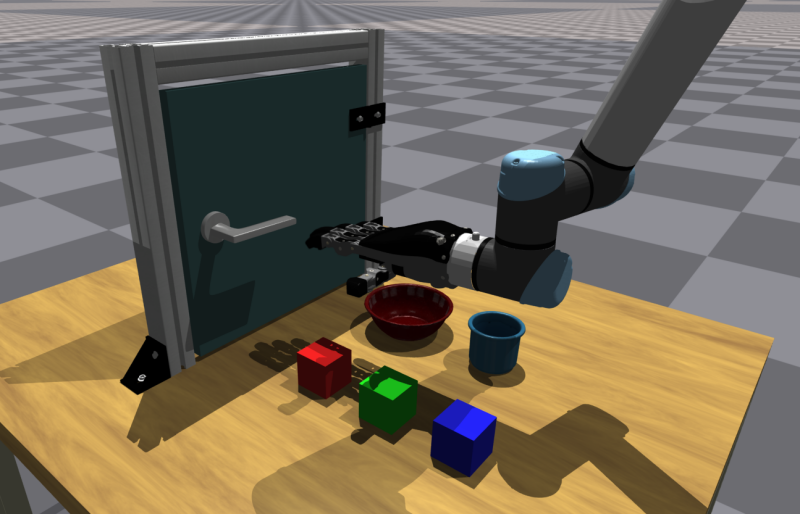}
      \caption{User study environment.}
      \label{user_study_environment}
\end{figure}

\subsection{Experimental Setup}
To answer the proposed questions, we selected a set of challenging manipulation tasks that mimic real-world problems that a humanoid robot might face (see Fig.~\ref{tasks}). The difficulty of these tasks depends on various factors. OpenDrawer is the simplest task, as it requires only a single skill and can be solved without very precise control of the fingers. OpenDoor is more complex, requiring the concatenation of the two skills of turning the handle and then pulling the door open. PourCup requires delicate manipulation and has a high risk of failure, since there is no way to recover if the cup is spilled somewhere other than the target location. The liquid in a real cup is represented by spherical particles, since we are working with a rigid-body physics simulation. This increased number of simulated bodies and contacts accounts for the lower simulation throughput on this task compared to the other environments (see Table~\ref{simulation_throughput}). The LiftObject task requires precise positioning of the fingers to produce a stable grip, as well as generalization to random initial positions and orientations of the object to be lifted.

The neural network architecture is kept constant across all evaluated methods. We represent the policy by an MLP consisting of 3 hidden layers of 512, 256, 256 neurons. Each RL agent is trained for 1000 epochs, where each epoch consists of 32 steps in the 16,384 parallel environment instances, resulting in a total 524 million steps.

\subsection{Haptic Teleoperation User Study}

First, we conduct a small user study to evaluate our teleoperation system. We analyze whether the system can be used by operators without prior experience to solve manipulation tasks. Further, we measure the influence of haptic feedback on user preference and task completion. Six participants with no prior experience operating the framework took part in our study. 

The participants were asked to perform three tasks in the environment shown in Figure~\ref{user_study_environment}. Specifically, the tasks were to stack three cubes, open a door, and pour particles from a cup into a bowl. A task was marked as a failure if a maximum time of 3\,min was reached or a non-recoverable failure occurred, e.g. dropping a cube off the table or spilling the cup. Each participant was asked to perform the tasks with and without haptic feedback. The order of these two trials was randomized to mitigate the influence of learning during teleoperation. 

Quantitative results of the user study in terms of success rate on the tasks and the time required are shown in Table~\ref{user_study}. During all trials, only a single failure occurred, where a cube fell off the table. The high success rates for diverse tasks show that the system is intuitive to use even without prior experience. Lower completion times were observed in all tasks when haptic feedback was enabled, highlighting the added value of integrating this modality into teleoperation. The task of stacking three cubes shows the highest standard deviation in completion time, since the tower can be knocked over during construction, requiring the user to start over.

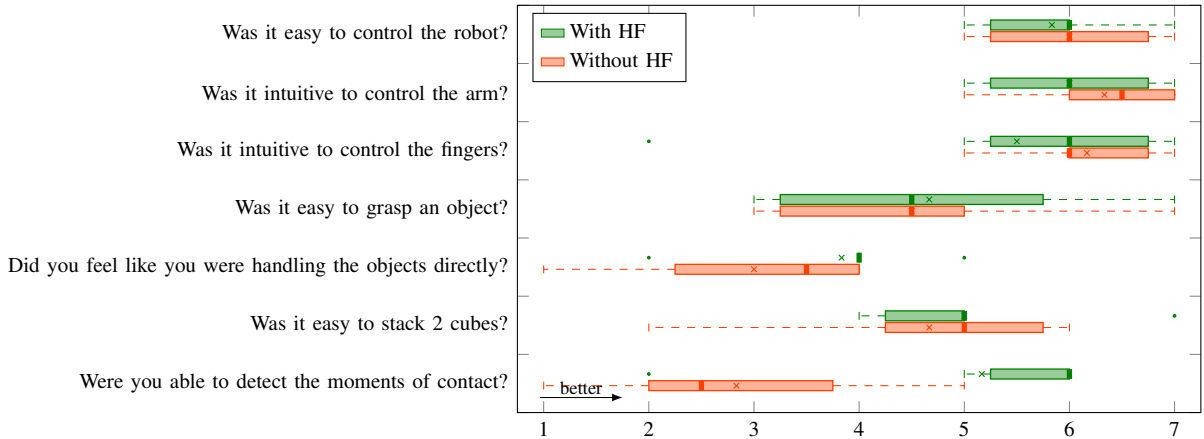
\begin{figure*}[t]
      \centering
      \input{user_study.pgf}
      \caption{Results of the user questionnaire. We depict the median, lower and upper quartile, lower and upper fence, outliers (•), and the average value ($\times$).}
      \label{user_questionnaire}
\end{figure*}

After each trial, participants were asked to rate their experience using seven-level Likert items. The results, shown in Fig.~\ref{user_questionnaire}, show that intuitive control of the robot arm and fingers was generally rated highly. Haptic feedback had a positive impact on whether users felt they were interacting directly with the objects. However, the most pronounced difference between the two modes of operation is evident in the ability to recognize moments of contact. Users reported that this feature of haptic feedback provided a more confident sense of the exact position of the hand relative to other objects.

\begin{table}[t]
\caption{Success rates and completion times for the tasks performed in the user study.}
\label{user_study}
\centering
\begin{tabular}{@{}crrrr@{}}\toprule
\multicolumn{1}{c}{\multirow{2}{4em}{Haptic feedback}} & \multicolumn{1}{c}{Task} & \multicolumn{1}{c}{Success rate} & \multicolumn{2}{c}{Completion time [s]} \\
\cmidrule{4-5}
&  &  & Mean & StdDev \\ \midrule
\multicolumn{1}{c}{\multirow{3}{4em}{\ding{51}}} & Stack cubes & 0.83 & \textbf{35.17} & \textbf{9.98} \\
& Open door & \textbf{1.0} & \textbf{6.47} & \textbf{1.82} \\
& Pour cup & \textbf{1.0} & \textbf{13.86} & \textbf{3.46} \\

\midrule

\multicolumn{1}{c}{\multirow{3}{4em}{\ding{55}}} & Stack cubes & \textbf{1.0} & 43.26 & 22.13 \\
& Open door & \textbf{1.0} & 9.37 & 5.26 \\
& Pour cup & \textbf{1.0} & 15.63 & 3.56 \\
\bottomrule
\end{tabular}
\end{table}

\subsection{Results and Analysis}

\begin{table*}[b]
\caption{Success Rates of evaluated methods. We performed at least three runs with 100 test episodes per run of each configuration.}
\label{success_rates}
\centering
\begin{tabular}{@{}rcrrrr@{}}\toprule
\multicolumn{1}{c}{Method} & \phantom{a} & \multicolumn{1}{c}{OpenDrawer} & \multicolumn{1}{c}{OpenDoor} & \multicolumn{1}{c}{PourCup} & \multicolumn{1}{c}{LiftObject} \\ \midrule
BC & &           $\textbf{1.0} \pm \textbf{0.0}$ & $0.96 \pm 0.02$ & $0.76 \pm 0.23$ & $0.27 \pm 0.03$ \\
PPO-dense & &    $\textbf{1.0} \pm \textbf{0.0}$ & $\textbf{1.0} \pm \textbf{0.0}$ & $0.98 \pm 0.01$ & $0.97 \pm 0.08$ \\
PPO-sparse & &   $\textbf{1.0} \pm \textbf{0.0}$ & $0.0 \pm 0.0$ & $0.48 \pm 0.48$ & $0.14 \pm 0.33$ \\
DAPG-sparse & &  $\textbf{1.0} \pm \textbf{0.0}$ & $\textbf{1.0} \pm \textbf{0.0}$ & $\textbf{1.0} \pm \textbf{0.0}$ & $\textbf{1.0} \pm \textbf{0.0}$ \\
\bottomrule
\end{tabular}
\end{table*}

In the following, we structure our analysis of the obtained results by the questions posed at the beginning of this section.

\begin{enumerate}
  \item[1)] \textit{Can the demonstrations collected with our VR teleoperation system be used to train successful policies for dexterous manipulation tasks?}
\end{enumerate}

To assess this question, we collected datasets of 200 demonstrations for each of the investigated tasks, which are split into 90\% training and 10\% validation data. We then train simple behavior cloning policies to replicate the observed demonstrations and use the validation loss to check for overfitting. The success rates are shown in the first row of Table~\ref{success_rates}. Near perfect results can be achieved using only imitation learning on the OpenDrawer and OpenDoor task. For PourCup, the results are still strong considering the complexity of the task and control space of the robot, but performance is not as consistent. Lastly, despite using only a single object geometry in this study, LiftObject is the most difficult to learn from demonstrations alone. This may be due to the fact that even for a human operator it is not straightforward to grasp an object securely, resulting in frequent regrasps or adjustments. Further, the generalization demanded by the random initialization of object poses appears to make imitation learning substantially harder. Overall, we were able to verify that the proposed pipeline can be used to collect demonstration for an anthropomorphic robot hand, but note that behavior cloning alone struggles to produce robust policies for more involved tasks. 

\begin{enumerate}
  \item[2)] \textit{How does learning from scratch in the massively parallel regime perform on the selected tasks?}
\end{enumerate}

We divide our analysis into the dense and sparse rewards setting. While it is generally more difficult to learn from sparse rewards, they are easy to specify. Providing dense rewards, on the other hand, requires tedious reward shaping and can bias learned behaviors towards unintended solutions. The strong results of PPO for dense rewards across all tasks studied suggest that massively parallel on-policy RL will find proficient solutions to tasks of the studied difficulty as long as a meaningful learning signal is present.
In contrast, sparse rewards were only sufficient to get satisfactory solutions to the OpenDrawer task. While LiftObject and PourCup make some progress towards solving the respective problem, the results vary strongly between seeds. OpenDoor made no learning progress across all seeds. In summary, sparse rewards were sufficient for learning only in simpler tasks where the solution can be discovered repeatedly through random exploration. Tasks that require the combination of multiple behaviors in succession along with precise actions are very difficult to solve in this way, even with the vast amounts of policy experience used in this study. 
\begin{enumerate}
  \item[3)] \textit{How does incorporating demonstration data into the learning process impact the performance of on-policy RL?}
\end{enumerate}

Finally, we analyze whether the integration of demonstrations into the learning process is sufficient to provide RL with the learning signal needed to generate robust policies. The perfect results for all tasks reported in the bottom of Table \ref{success_rates} confirm this hypothesis and highlight how the complementary strengths of massively parallel RL and imitation learning provide a powerful tool for solving challenging manipulation tasks. In this way, on-policy training in GPU-accelerated environments can be used to refine behaviors for which imitation learning alone does not lead to the desired robustness or generalization. For all tasks of the investigated complexity, this paradigm was able to produce robust policies, leading to no observed cases of failure in the final test-rollouts.  

\begin{figure}[t]
    \raggedright
    \textit{PPO} 
    \vspace{0.1cm}
    
    \centering
    \begin{subfigure}[b]{0.115\textwidth}
     \centering
     \includegraphics[width=\textwidth]{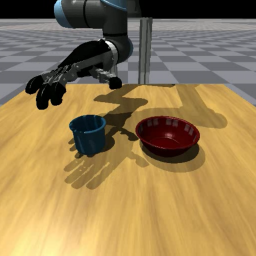}
    \end{subfigure}
    \hfill
    \begin{subfigure}[b]{0.115\textwidth}
     \centering
     \includegraphics[width=\textwidth]{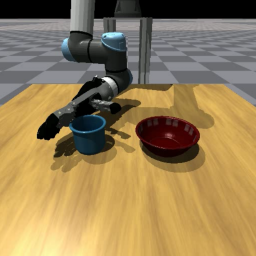}
    \end{subfigure}
    \hfill
    \begin{subfigure}[b]{0.115\textwidth}
     \centering
     \includegraphics[width=\textwidth]{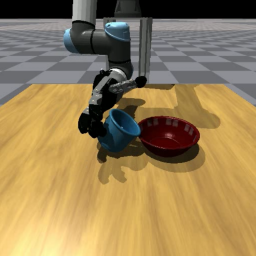}
    \end{subfigure}
    \hfill
    \begin{subfigure}[b]{0.115\textwidth}
     \centering
     \includegraphics[width=\textwidth]{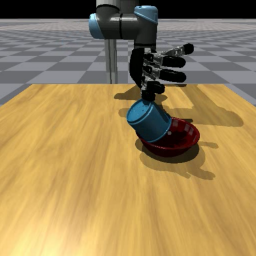}
    \end{subfigure}

    \raggedright
    \textit{DAPG} 
    \vspace{0.1cm}

     \begin{subfigure}[b]{0.115\textwidth}
         \centering
         \includegraphics[width=\textwidth]{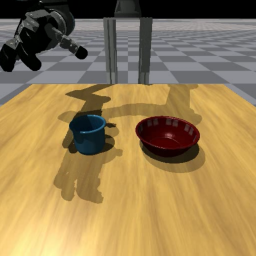}
     \end{subfigure}
     \hfill
     \begin{subfigure}[b]{0.115\textwidth}
         \centering
         \includegraphics[width=\textwidth]{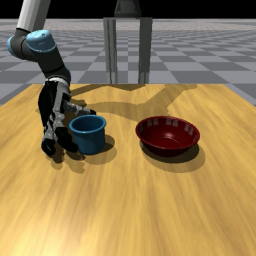}
     \end{subfigure}
     \hfill
     \begin{subfigure}[b]{0.115\textwidth}
         \centering
         \includegraphics[width=\textwidth]{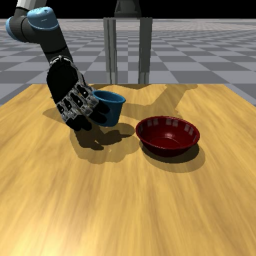}
     \end{subfigure}
     \hfill
     \begin{subfigure}[b]{0.115\textwidth}
         \centering
         \includegraphics[width=\textwidth]{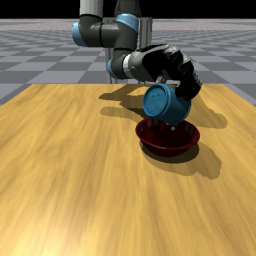}
     \end{subfigure}
      \caption{Different solution strategies employed by PPO and DAPG on the PourCup task.}
      \label{policy_rollouts}
\end{figure}

In Figure~\ref{policy_rollouts} we show examples of the learned behaviors for pure RL and DAPG on the PourCup task. An interesting difference is that PPO learns an unexpected strategy, where the particles in the cup are poured by knocking it over with the palm of the hand in just the right way. Even though this technically fulfills the objective of the problem, the intended behavior has not been learned. DAPG, on the other hand, is more likely to stick with the behaviors observed in the initial demonstrations. How much the algorithm deviates from these behaviors in favor of strategies that yield higher rewards is determined by the weighting terms $\lambda_0$ and $\lambda_1$.

\section{Discussion and Conclusions}
Our results show that adding demonstration data to the learning process of massively parallel model-free RL can leverage the complementary strengths of both approaches. Guidance from expert demonstrations provides RL with the training signal needed to make meaningful progress, while highly-parallelized RL training can be used to refine the behavior seen in the demonstrations. The purpose of this work is to highlight this useful connection and provide tools to foster research on the intersection of reinforcement and imitation learning. In future work, we aim to transfer policies learned in Nvidia Isaac Gym to a real robot setup. Therefore, accurate pose estimation or operation from visual perception directly will be required. Further interesting avenues for future work are to learn from demonstrations in multi-object environments, such as picking from cluttered bins, where generalization of the observed behaviors is crucial or to investigate whether a behavior cloning loss can be used to bias the solution found by RL, for example for functional grasps of objects. Lastly, our results showed that DAPG on GPU-accelerated RL environments is capable of solving all the manipulation tasks proposed here. Therefore, it would be interesting to investigate more complex multi-step tasks, such as removing an object from an initially closed drawer, to better explore the limitations of this approach.

\addtolength{\textheight}{-12cm}   




\printbibliography

\end{document}

%% file: user_study.pgf
\begin{tikzpicture}[font = \footnotesize, every mark/.append style={mark size=0.5pt}]
 \begin{axis}[
     name=plot,
     boxplot/draw direction=x,
     width=0.6\textwidth,
     height=7cm,
     boxplot={
         draw position={1.5 - 0.325/2 + 1.0*floor((\plotnumofactualtype + 0.001)/2) + 0.2*mod((\plotnumofactualtype + 0.001),2)},
         %
         box extend=0.17,
         average=auto,
         every average/.style={/tikz/mark=x, mark size=1.5, mark options=black},
         every box/.style={draw, line width=0.5pt, fill=.!40!white},
         every median/.style={line width=2.0pt},
         every whisker/.style={dashed},
     },
     ymin=1,
     ymax=8,
     y dir=reverse,
     ytick={1,2,...,9},
     y tick label as interval,
     yticklabels={
Was it easy to control the robot?,
Was it intuitive to control the arm?,
Was it intuitive to control the fingers?,
Was it easy to grasp an object?,
Did you feel like you were handling the objects directly?,
Was it easy to stack 2 cubes?,
Were you able to detect the moments of contact?,
     },
     y tick label style={
         align=center
     },
     xmin=0.75,
     xmax=7.25,
     xtick={1, 2 ,..., 7},
     xticklabels = {1, 2, ..., 7},
     cycle list={{green!50!black,orange!50!red}},
     y dir=reverse,
     legend image code/.code={
         \draw [#1, fill=.!40!white] (0cm,-1.5pt) rectangle (0.3cm,1.5pt);
     },
     legend style={
         anchor=north west,
         at={($(0.0,1.0)+(0.2cm,-0.1cm)$)},
     },
     legend cell align={left},
 ]


 \addplot
 table[row sep=\\,y index=0] {
 data\\
 6\\6\\5\\6\\7\\5\\
 };

 \addplot
 table[row sep=\\,y index=0] {
 data\\
 6\\5\\7\\6\\7\\5\\
 };


 \addplot
 table[row sep=\\,y index=0] {
 data\\
 5\\7\\6\\5\\7\\6\\
 };

 \addplot
 table[row sep=\\,y index=0] {
 data\\
 5\\7\\6\\7\\7\\6\\
 };


 \addplot
 table[row sep=\\,y index=0] {
 data\\
 6\\7\\2\\6\\7\\5\\
 };

 \addplot
 table[row sep=\\,y index=0] {
 data\\
 6\\7\\6\\6\\7\\5\\
 };


 \addplot
 table[row sep=\\,y index=0] {
 data\\
 4\\7\\6\\3\\3\\5\\
 };

 \addplot
 table[row sep=\\,y index=0] {
 data\\
 5\\7\\4\\5\\3\\3\\
 };


 \addplot
 table[row sep=\\,y index=0] {
 data\\
 4\\2\\4\\5\\4\\4\\
 };

 \addplot
 table[row sep=\\,y index=0] {
 data\\
 3\\1\\4\\4\\4\\2\\
 };


 \addplot
 table[row sep=\\,y index=0] {
 data\\
 5\\7\\5\\4\\4\\5\\
 };

 \addplot
 table[row sep=\\,y index=0] {
 data\\
 5\\6\\5\\6\\2\\4\\
 };


 \addplot
 table[row sep=\\,y index=0] {
 data\\
 6\\2\\6\\6\\6\\5\\
 };

 \addplot
 table[row sep=\\,y index=0] {
 data\\
 2\\1\\4\\3\\5\\2\\
 };

 \legend{With HF, Without HF}

 \end{axis}

 \draw[-latex] ($(plot.south west)+(0.3cm,0.2cm)$) -- ($(plot.south west)+(1.4cm,0.2cm)$)  node[midway,above,font=\scriptsize,inner sep=1pt] {better};

 \end{tikzpicture}